\documentclass[final,twocolumn,9pt]{IEEEtran}

\usepackage{spconf,graphicx}

\usepackage{amsmath,epsfig}

\usepackage{comment}
\usepackage{url} 
\usepackage{hyperref}
\usepackage{graphics}

\usepackage{multirow}

\title{Dense-Haze: A benchmark for image dehazing with dense-haze and haze-free images}

\name{Codruta O. Ancuti$^{*}$, Cosmin Ancuti $^{*}$, Mateu Sbert$^{\dag}$, Radu Timofte$^{\ddag}$}

\address{$^{*}$  ETcTI, Universitatea Politehnica Timisoara, Romania\\
$^{\dag}$ Institute of Informatics and Applications, University of Girona, Spain\\
$^{\ddag}$ CVL, ETH Zurich, Switzerland \\
}

\begin{document}
\maketitle
\begin{abstract}
Single image dehazing is an ill-posed problem that has recently drawn important attention. Despite the significant increase in interest shown for dehazing over the past few years, the validation of the dehazing methods remains largely unsatisfactory, due to the lack of pairs of real hazy and corresponding haze-free reference images.
To address this limitation, we introduce \textbf{Dense-Haze} -- a novel dehazing dataset. Characterized by dense and homogeneous hazy scenes, \textbf{Dense-Haze} contains $33$ pairs of real hazy and corresponding haze-free images of various outdoor scenes. The hazy scenes have been recorded by introducing real haze, generated by professional haze machines. The hazy and haze-free corresponding scenes contain the same visual content captured under the same illumination parameters. \textbf{Dense-Haze} dataset aims to push significantly the state-of-the-art in single-image dehazing by promoting robust methods for real and various hazy scenes. We also provide a comprehensive qualitative and quantitative evaluation of state-of-the-art single image dehazing techniques based on the \textbf{Dense-Haze} dataset. Not surprisingly, our study reveals that the existing dehazing techniques perform poorly for dense homogeneous hazy scenes and that there is still much room for improvement.
\end{abstract}

\section{Introduction}

Haze represents one of the atmospheric phenomena most challenging for camera sensors and vision applications. Haze is an often occurring meteorological phenomenon especially during autumn and spring in temperate climates being generated by small floating particles which absorb and scatter the light from its propagation direction. The  visibility of such hazy scene is highly degraded generating loss of contrast for the distant objects, selective attenuation of the light spectrum, and additional noise. For instance, the presence of haze has a great impact in the road traffic as it may severely reduce the  visibility for drivers. As a result, restoring the contents in hazy images -- process known as \textit{dehazing} -- is important for several outdoor image processing and computer vision applications such as visual surveillance and automatic driving assistance.

To solve this problem, earlier techniques employ polarization filters~\cite{PAMI_2003_Narasimhan_Nayar,Schechner_2003}, and depth knowledge prior~\cite{Kopf_DeepPhoto_SggAsia2008,Tarel_CVPR_2007}.  On the other hand, restoring the visibility in hazy images based only on a single input RGB image is more challenging, being a mathematically ill-posed problem~\cite{Fattal_Dehazing,Tan_Dehazing,Kratz_and_Nishino_2009,Tarel_ICCV_2009,Dehaze_He_CVPR_2009,Nishino_2012,Galdran_2018,Liu_dehazing_2018}. Fattal~\cite{Fattal_Dehazing} introduces a Markov Random Field (MRF) method to search for a solution in which the resulting shading and transmission functions are locally statistically uncorrelated.
He et al.~\cite{Dehaze_He_CVPR_2009} propose the Dark Channel Prior (DCP), a simple but effective solution to estimate the transmission map. Instead of using alpha-matting as in~\cite{Dehaze_He_CVPR_2009}, the method of Meng et al.~\cite{Meng_2013} is based on a regularization strategy and refines the boundaries of the rough transmission estimated by DCP. Similarly, Zhu et al.~\cite{Zhu_2015} extends DCP by considering a color attenuation prior under the assumption that the depth can be estimated from pixel saturation and intensity. The color-lines model~\cite{Fattal_Dehazing_TOG2014} has been extended recently by Berman et al.~\cite{Berman_2016} using the observation that the colors of a haze-free image are well approximated by a limited set of tight clusters in the RGB space. Another category is represented by multi-scale fusion approaches~\cite{Ancuti_TIP_2013,Choi_2015} that enhance the hazy scenes without explicitly estimating  the  transmission map. More recently, several machine learning dehazing methods~\cite{Tang_2014,Dehazenet_2016,Ren_2016,Zhang_dehazing_2018,Santra_2018,Liu_2019,Wang_2019}  have been introduced in the literature. DehazeNet~\cite{Dehazenet_2016} is trained based on a synthetically built dehazing dataset to estimate the transmission map which is subsequently used to compute a haze-free image via traditional optical model. Employing also a synthesized dataset in the training stage, Ren et al.~\cite{Ren_2016} define  a coarse-to-fine neural network consisting of a cascade of CNN layers.

\begin{figure}[t!]
  \centering
  \includegraphics[width=1\linewidth]{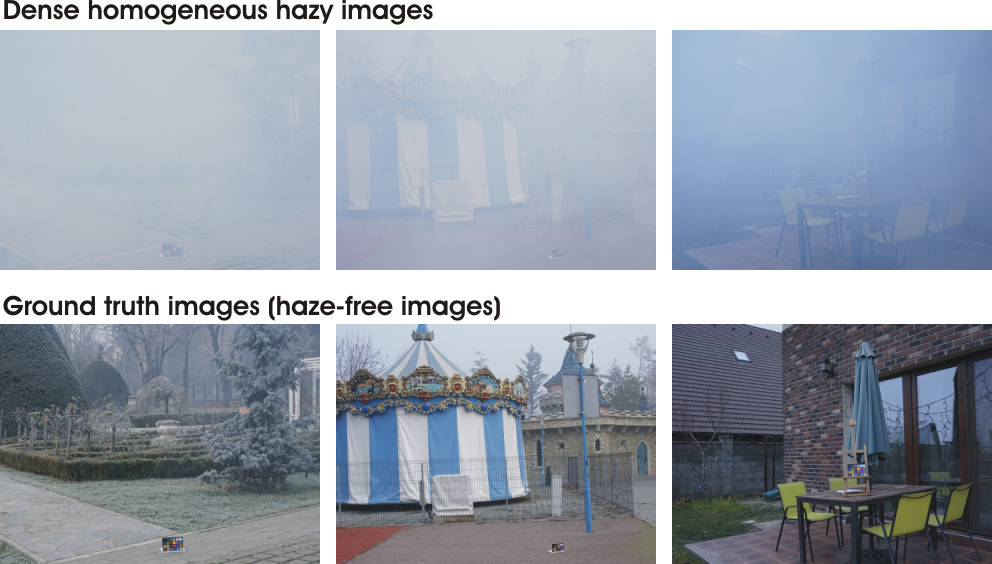}
  \caption{\label{fig:intro}%
     \textit{Three examples of the \textbf{Dense-Haze} dataset that provides $33$ pairs of hazy and corresponding haze-free, i.e. groundtruth, outdoor images.} 
  }  
\end{figure}

Although the interest shown for the image dehazing problem has increased significantly over the past few years, the validation of the proposed dehazing methods remain largely unsatisfactory, mainly due to the absence of pairs of corresponding hazy and haze-free ground-truth images.  Recording realistic images is very challenging and time-consuming due to the practical issues associated to the recording of reference and hazy images under identical illumination condition. As a result, most of the existing dehazing assessment datasets~\cite{Tarel_2012,D_Hazy_2016,HazeRD_2017} rely on synthesized hazy images using a simplified optical model and known depth. Tarel et al.~\cite{Tarel_2012} introduced the first synthetic dehazing dataset (FRIDA) which contains 66 computer graphics generated traffic scenes. D-HAZY~\cite{D_Hazy_2016} is another synthetic dehazing dataset with 1400+ real images and corresponding depth maps used to synthesize hazy scenes based on Koschmieder's light propagation model~\cite{Koschmieder_1924}. HazeRD~\cite{HazeRD_2017} extends D-HAZY dataset by adding several synthesized outdoor hazy images. Luthen et al.~\cite{RGB_NIR_2017} introduced a dataset consisting of four sets of indoor scenes with hazy and RGB and NIR ground-truth images. 

So far the focus in the dehazing literature has been on relatively light hazy conditions which potentially limits the utility of the proposed existing dehazing techniques for real scenes with dense haze. The introduction of a dehazing dataset with dense haze and corresponding haze-free reference images is very important to assess the existing dehazing techniques and furthermore to advance the research in the dehazing field.

The main contribution of this paper is \textbf{Dense-Haze}, a new realistic dehazing dataset. Characterized by dense and homogeneous hazy scenes, \textbf{Dense-Haze} contains $33$ pairs of real hazy and corresponding haze-free images of various outdoor scenes. In order to generate hazy scenes we used a professional haze machine that imitates with high fidelity real haze. To preserve the illumination conditions, all the outdoor scenes are static and have been recorded during cloudy days, in the morning or in the sunset. Basically, \textbf{Dense-Haze} extends the O-HAZE~\cite{Ancuti_OHAZE_2018} dataset that has been used recently for the first single image dehazing challenge ever organized~\cite{Ancuti_NTIRE_2018}. In contrast to O-HAZE that contains only light hazy scenes, \textbf{Dense-Haze} is more challenging since all the recorded scenes contain a denser and more homogeneous haze layer (see Fig.~\ref{fig:intro}). We believe that introducing the \textbf{Dense-Haze} dataset will push significantly the state-of-the-art in single-image dehazing methods making them to be more robust for real and various dense haze scenes.

A second contribution of this paper is a comprehensive qualitative and quantitative evaluation of the state-of-the-art single image dehazing techniques based on the \textbf{Dense-Haze} dataset. In our study we compare a set of representative dehazing methods and evaluate them using traditional measures such as PSNR and SSIM on \textbf{Dense-Haze} dataset. 
Our experimental results reveal that the existing dehazing techniques perform poorly for dense hazy scenes, which was somewhat expected given the fact the most of the existing methods were introduced and validated on lighter haze conditions. There is clearly much room for improvement and our proposed \textbf{Dense-Haze} dataset can promote and benchmark research for robust image dehazing solutions.

\section{Recording \textbf{Dense-Haze} dataset}

In this section we discuss the methodology of recording the $33$ pairs of hazy and haze-free (ground-truth) outdoor images of the \textbf{Dense-Haze} dataset. As we briefly discussed, a crucial problem in collecting such images is represented by capturing pixel-level images for each scene with and without haze under identical conditions, using the same camera settings, viewpoint, etc. Besides assuring that the scene is static, the scene components keep do not change their spatial position during the recording (quite challenging for natural scenes due to numerous factors), the most challenging issue is to preserve the outdoor scene illumination.

As a result, we recorded the outdoor scenes only during cloudy days, in the morning or in the sunset. Additionally, another important constraint was given by the influence of the wind. In order to limit fast spreading of the haze in the scene we could record images only when the wind speed was below 2-3 km/h. This constraint was hard to meet, it is a main reason for the 8 weeks duration required by the recording of the 33 outdoor scenes from \textbf{Dense-Haze}. 

To yield hazy scenes, the haze was spread using two professional haze  machines (LSM1500 PRO 1500 W), which generate vapor particles with diameter size (typically 1 - 10 microns) similar to the particles of the atmospheric haze. The haze machines use cast or platen type aluminum heat exchangers to induce liquid evaporation. In order to simulate the effect occurring with water haze over larger distances than the investigated 20-30 meters, we used special (haze) liquid with higher density. To obtain a dense and homogeneous haze layer in the scene, we employed for 2-3 minutes both haze machines powered by a portable 2800 Watt generator, and waited for another 2-3 minutes.

Before introducing haze in the scenes, we settled the recording setup composed by a tripod and a Sony A5000 camera that was remotely controlled (Sony RM-VPR1). This setup allowed to acquire JPG and ARW (RAW) $5456 \times 3632$ images, with 24 bit depth. For each scene recording a manual adjustment of the camera settings has been performed. We use the same  camera setting to capture the haze-free and hazy images of the same scene. Basically, the camera parameters related to the  shutter-speed (exposure-time), the aperture (F-stop), the ISO and white-balance have been kept identical when recording hazy and haze-free scenes. Therefore, the closer regions (that in general are less distorted by haze) have similar appearance (in terms of color) in the corresponding scenes.    

The optimal camera parameters (aperture-exposure-ISO), have been set based on the built-in light-meter of the camera, but also using an external exponometer (Sekonic). For the custom white-balance, we used the middle gray card (18\% gray) of the color checker. This is a common photographic process that requires to use the camera white-balance mode in manual mode and place the reference gray-card in the front of the camera (the gray-card was placed in the center of the scene in the range of four meters). Additionally, all the recorded scenes contain a color checker to allow for the post-processing of the recorded images. We used a classical Macbeth color checker with the size 11 by 8.25 inches with 24 squares of painted samples (4$\times$6 grid).

\section{Evaluated Dehazing Techniques}

\begin{figure*}[t!]
  \centering
  \includegraphics[width=0.86\linewidth]{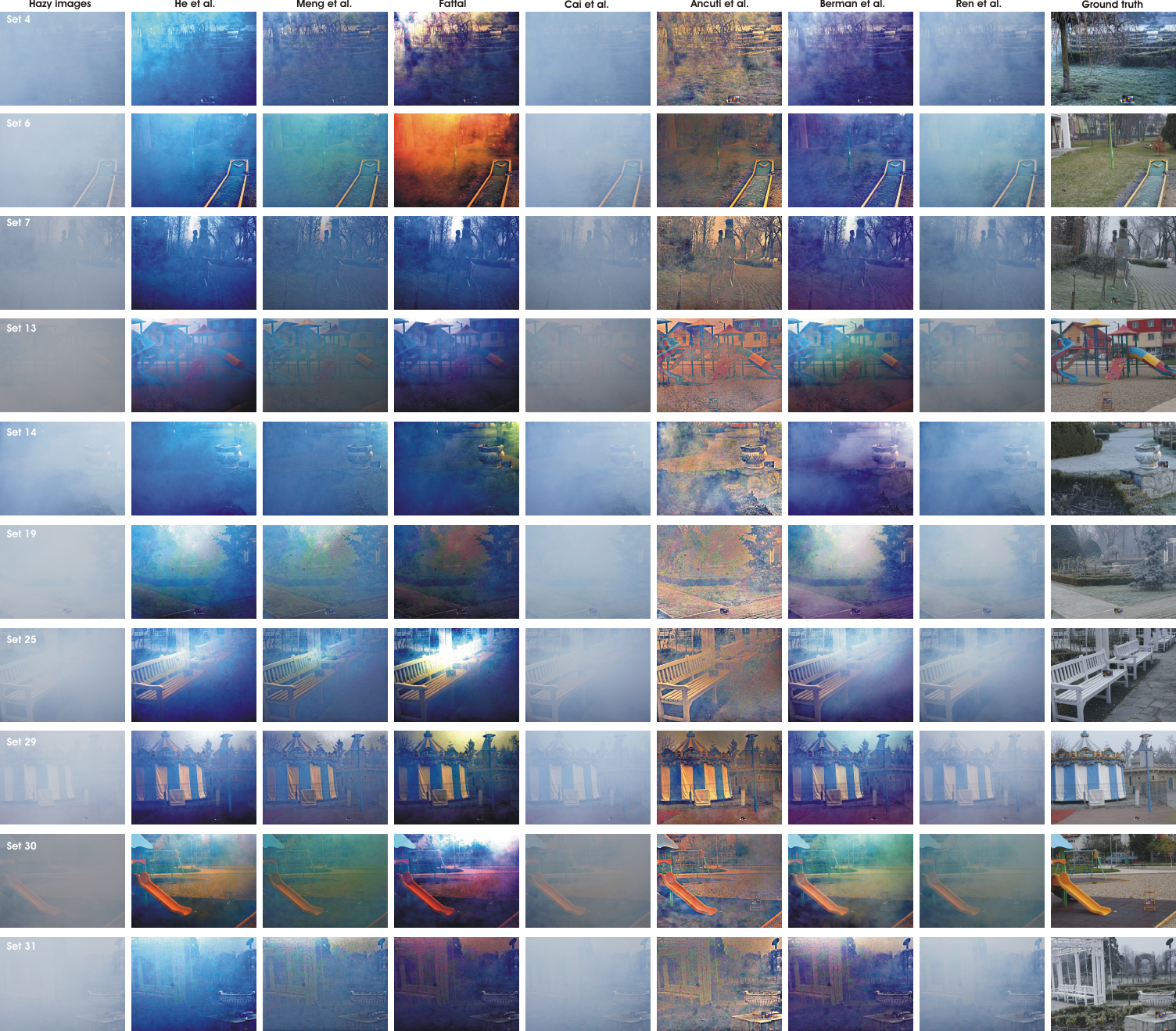}
  \caption{\label{fig:res_comp1}%
    \textit{\textbf{Comparative results of representative single image dehazing techniques.} The first row shows the hazy images and the last row shows  the ground truth. The middle rows, from left to right, present the results of He et al.~\cite{Dehaze_He_CVPR_2009}, Meng et al.~\cite{Meng_2013}, Fattal~\cite{Fattal_Dehazing_TOG2014}, Cai et al.~\cite{Dehazenet_2016}, Ancuti et al.~\cite{Ancuti_NightTime},  Berman et al.~\cite{Berman_2016} and Ren et al.~\cite{Ren_2016}.}
  }  
\end{figure*}

\begin{table*}[]
\centering
\begin{tabular}{|l|l|l|l|l|l|l|l|l|l|l|l|l|l|l|}
\hline
\multirow{2}{*}{} & \multicolumn{2}{l|}{\small{\textbf{He et al.}}~\cite{Dehaze_He_CVPR_2009}} & \multicolumn{2}{l|}{\small{\textbf{Meng et al.}}~\cite{Meng_2013}} & \multicolumn{2}{l|}{\small{\textbf{Fattal}}~\cite{Fattal_Dehazing_TOG2014}} & \multicolumn{2}{l|}{\small{\textbf{Cai et al.}}~\cite{Dehazenet_2016}} & \multicolumn{2}{l|}{\small{\textbf{Ancuti et al.}}~\cite{Ancuti_NightTime}} & \multicolumn{2}{l|}{\small{\textbf{Berman}}~\cite{Berman_2016}} & \multicolumn{2}{l|}{\small{\textbf{Ren et al.}}~\cite{Ren_2016}} \\ \cline{2-15} 
                  & {\scriptsize{PSNR}}             & {\tiny{CIEDE2000}}            & {\scriptsize{PSNR}}              & {\tiny{CIEDE2000}}            & {\scriptsize{PSNR}}            & {\tiny{CIEDE2000}}          & {\scriptsize{PSNR}}              & {\tiny{CIEDE2000}}            & {\scriptsize{PSNR}}               & {\tiny{CIEDE2000}}              & {\scriptsize{PSNR}}               & {\tiny{CIEDE2000}}              & {\scriptsize{PSNR}}              & {\tiny{CIEDE2000}}            \\ \hline
\textbf{Set 4 }            & 14.12        & 21.54           & 14.57         & 21.22            & 11.10      & 27.98          & 10.08        & 28.50            & 14.39          & 24.89             & 12.55          & 26.08             & 11.39        & 25.04            \\ \hline
\textbf{Set 6}   & 15.99        & 30.19           & 15.47         & 27.43            & 12.32      & 33.31          & 10.97        & 30.40            & 15.74          & 19.51             & 14.79          & 34.92             & 12.46        & 28.05            \\ \hline
\textbf{Set 7 }            & 15.65        & 22.23           & 16.97         & 21.13            & 14.59      & 26.84          & 13.95        & 20.30            & 18.48          & 17.07             & 15.29          & 26.80             & 16.45        & 18.66            \\ \hline
\textbf{Set 13}  & 13.43        & 24.04           & 15.03         & 21.85            & 11.64      & 30.74          & 14.54        & 19.50            & 17.57          & 19.84             & 12.98          & 25.85             & 15.34        & 18.88            \\ \hline
\textbf{Set 14}    & 15.06        & 24.63           & 14.67         & 24.40            & 13.11      & 27.08          & 9.58         & 30.00            & 11.18          & 28.31             & 12.62          & 28.02             & 11.78        & 25.59            \\ \hline
\textbf{Set 19}   & 14.42        & 20.27           & 13.92         & 22.08            & 10.71      & 27.24          & 11.09        & 24.44            & 13.27          & 24.52             & 12.24          & 24.84             & 11.29        & 23.35            \\ \hline
\textbf{Set 25}   & 14.51        & 19.88           & 14.78         & 20.72            & 11.54      & 26.28          & 11.20        & 22.06            & 15.62          & 21.00             & 11.60          & 24.89             & 12.29        & 20.06            \\ \hline
\textbf{Set 29}  & 14.36        & 19.72           & 14.86         & 19.53            & 11.06      & 28.15          & 14.00        & 18.06            & 14.77          & 19.08             & 13.99          & 22.85             & 16.00        & 14.85            \\ \hline
\textbf{Set 30}   & 14.69        & 19.78           & 15.12         & 23.23            & 10.48      & 33.03          & 13.45        & 20.13            & 16.24          & 20.72             & 13.72          & 22.47             & 14.58        & 20.55            \\ \hline
\textbf{Set 31 }           & 14.22        & 21.79           & 14.60         & 21.19            & 9.88       & 32.66          & 11.61        & 22.11            & 13.98          & 21.99             & 13.03          & 25.43             & 12.52        & 20.59            \\ \hline
\end{tabular}
\caption{\label{tabel_eval1} \textit{\textbf{Quantitative evaluation.} In this table are presented 10 randomly picked up  sets  from our \textbf{Dense-Haze} dataset (the hazy images, ground truth and the results are shown in Fig.\ref{fig:res_comp1}). Using the haze-free (ground-truth) images we can compute the PSNR and CIEDE2000 values for the  dehazed images produced by the evaluated techniques.}}
\end{table*}

\begin{table*}[h!]
\centering
\label{table_average}
\begin{tabular}{|l|l|l|l|l|l|l|l|}
\hline
          & {\small{\textbf{He et al.}}~\cite{Dehaze_He_CVPR_2009}} & {\small{\textbf{Meng et al.}}~\cite{Meng_2013}} & {\small{\textbf{Fattal}}~\cite{Fattal_Dehazing_TOG2014}} & {\small{\textbf{Cai et al.}}~\cite{Dehazenet_2016}} & {\small{\textbf{Ancuti et al.}}~\cite{Ancuti_NightTime}} & {\small{\textbf{Berman et al.}}~\cite{Berman_2016}} & {\small{\textbf{Ren et al.}}~\cite{Ren_2016}} \\ \hline
\small{\textbf{SSIM}}      & \textbf{0.398 }    & 0.352      & 0.326  & 0.374      & 0.306         & 0.358         & 0.369     \\ \hline
\small{\textbf{PSNR}}      & 14.557    & \textbf{14.621}     & 12.114 & 11.362     & 13.669        & 13.176        & 12.524    \\ \hline
\small{\textbf{CIEDE2000}} & \textbf{23.388 }   & 23.420     & 27.834 & 26.879     & 24.417        & 27.918        & 24.689     \\ \hline
\end{tabular}
\caption{Quantitative evaluation of the entire \textbf{Dense-Haze} dataset. This table presents the average values of the SSIM, PSNR and CIEDE2000 indexes, over the entire dataset (33 sets of images). }
\end{table*}

In this work we  evaluate qualitatively and quantitatively several state-of-the-art  single image dehazing techniques based on the  \textbf{Dense-Haze} dataset. For the sake of completeness,  in the following paragraphs we briefly review the evaluated dehazing methods.
\newline
\newline
\textbf{He et al.}~\cite{Dehaze_He_CVPR_2009}, introduce probably the most influential single-image dehazing approach. The authors of~\cite{Dehaze_He_CVPR_2009}  define the Dark Channel Prior (DCP),  a statistic observation  that yields a rough  estimate (per patch) of the transmission map. DCP is an  heuristic approach that  builds on the observation that most of the local regions (with the exception of the sky or hazy regions) contain pixels that present low intensity in at least one of the color channels. The roughly estimated transmission  is  refined by an  alpha matting strategy~\cite{Dehaze_He_CVPR_2009} or by using guided filter~\cite{Guided_filter_PAMI_2013}. In our evaluation, the results of \textbf{He et al.}~\cite{Dehaze_He_CVPR_2009} have been generated using the DCP refined with guided filters.\\   
\newline
\textbf{Meng et al.}~\cite{Meng_2013} introduce a method that builds on the DCP~\cite{Dehaze_He_CVPR_2009}. The estimated transmission map  (based on DCP) is refined further using  a boundary constraint that  is  combined with a weighted L1−norm regularization. Overall, this method  mitigates the lack of resolution in the DCP transmission map. Moreover, the method of Meng  et al.~\cite{Meng_2013} demonstrates some improvement compared with the He et al.~\cite{Dehaze_He_CVPR_2009} technique, reducing the  artifacts level close to the sharp edges.\\
\newline
\textbf{Fattal}~\cite{Fattal_Dehazing_TOG2014} makes use of color-lines in the RGB color space firstly introduced by Omer et al.~\cite{Omer_2004}.  The method is built on the  observation that the distributions of pixels in small natural image patches exhibit one-dimensional structures. This finding allows to compute a rough estimate of the transmission map that is further refined by employing a Markov Random Field model that filters the noise and removes some artifacts due to the scattering. \\
\newline
\textbf{Cai et al.}~\cite{Dehazenet_2016}  introduces \textbf{DehazeNet}, a convolutional neural network (CNN) approach that trains a model to map hazy to haze-free patches. \textbf{DehazeNet}  is characterized by  four sequential steps: features extraction, multi-scale mapping, local extrema and finally non-linear regression. The model is trained using  a synthesized dehazing dataset. \\
\newline
\textbf{Ancuti et al.}~\cite{Ancuti_NightTime} rely  also on DCP, but their work the authors  introduce a simple method to estimate  locally the airlight constant. Deriving several input images obtained from distinct definitions of the locality notion, the method  employs a  multi-scale fusion strategy. Designed initially as a solution for more  complex night-time hazy scenes (that are characterized by severe scattering and multiple sources of light), this fusion-based strategy shown to be competitive also for day-time single-image dehazing. \\ 
\newline
\textbf{Berman et al.}~\cite{Berman_2016} extends the color-lines concept of~\cite{Fattal_Dehazing_TOG2014} considering  that the color distribution in a haze-free images is well approximated by a discrete set of clusters in the RGB  color space. The method builds on the observation that  the pixels in a given cluster are non-local and are spread over the entire image plane. As a result, the pixels in a cluster are affected differently by the haze and convey information that can be used to estimate the transmission map.\\ 
\newline
\textbf{Ren et al.}~\cite{Ren_2016} is also a a convolutional neural network (CNN) strategy, but different than~\cite{Dehazenet_2016}, the transmission map is firstly computed by a coarse-scale network, and  then iti is refined by a fine-scale network. Similarly,  the training of the network is based on a synthetically generated dehazing dataset, obtained from haze-free images and their associated depth maps  employed as a transmission map in the  simplified  optical model. 
\newline

\section{Evaluation and Discussion}

The $33$ pairs of hazy and haze-free (ground-truth) outdoor images of the \textbf{Dense-Haze} dataset have been used to evaluate several representative single image dehazing techniques that were briefly described in the previous section. In Fig.\ref{fig:res_comp1} are shown 7 scenes of the \textbf{Dense-Haze} dataset and the dehazed image results generated using the methods of He et al.~\cite{Dehaze_He_CVPR_2009}, Meng et al.~\cite{Meng_2013}, Fattal~\cite{Fattal_Dehazing_TOG2014}, Cai et al.~\cite{Dehazenet_2016}, Ancuti et al.~\cite{Ancuti_NightTime},  Berman et al.~\cite{Berman_2016} and Ren et al.~\cite{Ren_2016}.

By analyzing the visual results presented in Fig.\ref{fig:res_comp1}, we can observe that in general the DCP-based techniques~\cite{Dehaze_He_CVPR_2009,Meng_2013,Ancuti_NightTime} recover the global image structure, but introduce unpleasing color shifting in the hazy regions mostly due to the poor airlight estimation. These distortions are more significant in the lighter/whiter regions, where the dark channel prior usually fails. Similarly, the methods of  Fattal~\cite{Fattal_Dehazing_TOG2014} and Berman et al~\cite{Berman_2016} introduce unpleasing color artifacts.
Not surprisingly, although they do not introduce additional distortions, the learning-based approaches of Ren et al~\cite{Ren_2016} and Cai et al~\cite{Dehazenet_2016}, trained using synthetic hazy dataset, are not able to remove the hazy (white) appearance. Overall, Fig.~\ref{fig:res_comp1} demonstrates that all the single image dehazing techniques from this study perform quite poorly for scenes from the \textbf{Dense-Haze} dataset. We conclude that the analyzed methods introduce structural distortions close to the sharp transitions with the artifacts more visible in regions far away from the camera. Moreover, in some cases the color distortions of the dehazed results look unnatural.

In addition to qualitative evaluation, \textbf{Dense-Haze} dataset makes possible for an objective quantitative evaluation of the single-image dehazing techniques. The haze-free (ground-truth) images available in our dataset allow to evaluate the quality of the corresponding dehazed results as a per-pixel fidelity to the ground-truth. In our evaluation,  we considered the Peak Signal-to-Noise Ratio
(PSNR), Structure Similarity Index Measure (SSIM)~\cite{Wang_2004} and CIEDE2000~\cite{Sharma_2005,Westland_2012}. While PSNR measures absolute
errors, SSIM is a perception-based model that considers local patterns of pixel intensities that have been normalized for luminance and contrast. SSIM assesses results in the ranges in [-1,1], with maximum value 1 for two identical images. For color appearance, we  consider  CIEDE2000~\cite{Sharma_2005,Westland_2012} that measures the color difference between two images and generates smaller values for better color preservation.

The quantitative results based on PSNR and CIEDE2000 of the image pairs shown in Fig.~\ref{fig:res_comp1} are reported in Table~\ref{tabel_eval1}. Moreover, in Table~\ref{table_average} we summarize the overall quantitative results over the entire \textbf{Dense-Haze} dataset. From these tables, we can conclude that the He et al.~\cite{Dehaze_He_CVPR_2009} and Meng et al.~\cite{Meng_2013}  perform slightly better in terms of structure and color restoration  compared with the other techniques. 
Overall, all analyzed dehazing methods achieve a relatively low performance in SSIM, PSNR and CIEDE2000 terms. This demonstrates once again the complexity of the image dehazing problem. 

\textbf{Conclusions:} We introduce a novel dehazing dataset that contains dense-hazy scenes and their counterpart haze-free images. As revealed,  by our proposed \textbf{Dense-Haze} dataset, the existing image dehazing techniques are not prepared to deal with dense hazy scenes and leaves significant room for improvement both qualitatively and quantitatively.\newline

\bibliographystyle{IEEEbib}
\bibliography{ref}

\end{document}